\pdfoutput=1

\documentclass[11pt]{article}

\usepackage[]{ACL2023}

\usepackage{times}
\usepackage{latexsym}
\usepackage{booktabs}
\usepackage{enumitem}
\usepackage[nameinlink]{cleveref}

\usepackage[T1]{fontenc}

\usepackage[utf8]{inputenc}

\usepackage{microtype}

\usepackage{inconsolata}

\title{RakutenAI-7B: Extending Large Language Models for Japanese}

\author{{Rakuten Group, Inc.}\thanks{{ }{ }The authors are listed in the alphabetical order by their first name.} \\\AND \small
Aaron Levine, Connie Huang, Chenguang Wang, Eduardo Batista, Ewa Szymanska, Hongyi Ding, \\\AND \small
Hou Wei Chou, Jean-François Pessiot, Johanes Effendi, Justin Chiu, Kai Torben Ohlhus,\\\AND \small
Karan Chopra, Keiji Shinzato, Koji Murakami, Lee Xiong, Lei Chen, Maki Kubota, Maksim Tkachenko,\\\AND \small
Miroku Lee, Naoki Takahashi, Prathyusha Jwalapuram, Ryutaro Tatsushima, Saurabh Jain, \\\AND \small
Sunil Kumar Yadav, Ting Cai, Wei-Te Chen, Yandi Xia, Yuki Nakayama, Yutaka Higashiyama}

\begin{document}

\maketitle

\begin{abstract}
We introduce RakutenAI-7B, a suite of Japanese-oriented large language models that achieve the best performance on the Japanese LM Harness benchmarks among the open 7B models. Along with the foundation model, we release instruction- and chat-tuned models, RakutenAI-7B-instruct and RakutenAI-7B-chat respectively, under the Apache 2.0 license. 

\end{abstract}
\section{Introduction}
\label{sec:intro}

The shift towards a new ``Pre-Train, Prompt, and Predict'' paradigm \cite{liu2021pretrain} in natural language processing (NLP) has resulted in the active development of large language models (LLMs). These models deliver a unified solution to a wide variety of NLP tasks, often showcasing a superior performance over the ``Pre-Train and Fine-Tune'' paradigm. This new research trend, however, has primarily been focused on English leaving out languages such as Japanese.

RakutenAI-7B is a systematic initiative that brings the latest technologies to the world of Japanese LLMs. RakutenAI-7B achieves the best scores on the Japanese language understanding benchmarks while maintaining a competitive performance on the English test sets among similar models such as OpenCalm \cite{OpenCalm7b}, Elyza \cite{elyzallama2023}, Youri \cite{RinnaYouri7b}, Nekomata \cite{RinnaNekomata7b}, and Swallow \cite{swallow7b}. RakutenAI-7B leverages the Mistral model architecture \cite{jiang2023mistral} and is based on Mistral 7B-v0.1 pre-trained checkpoint, exemplifying a successful retrofitting of the pre-trained model weights. Moreover, we extend Mistral's vocabulary from 32k to 48k to offer a better character-per-token rate for Japanese.

RakutenAI-7B achieves a significant milestone for Japanese-oriented LLMs, while maintaining a competitive English performance. Our aim is to help the community to create more affordable and efficient Japanese language models that can be used in a wide variety of applications. We release our models to the public under the Apache 2.0 License. The models are accessible at \url{https://huggingface.co/Rakuten/RakutenAI-7B}. 

In the rest of the report, we describe the key aspects of RakutenAI-7B, delving into the tokenizer extension, pre-training, fine-tuning, and model evaluation. 

\section{Technical Details}
\label{sec: technical_details}

\subsection{Tokenization}
\label{ssec: tokenization}

The Mistral tokenizer often encodes a single Japanese character into multiple tokens, which has two implications. First, as the language model operates within a limited context size, it significantly reduces the length of Japanese text that can be processed by the model. Second, text generation with the low character-per-token rate tokenizer would spend more GPU cycles to generate the same text, as opposed to a tokenizer with a higher rate. The problem stems from the logographic part of Japanese (kanji): Mistral's limited vocabulary size is not optimized specifically to parse the kanji characters, which causes the tokenizer to encode a large proportion of them as a sequence of bytes. This results in a character-per-token rate that is less than 1. The character-per-token rate is the ratio between the number of UTF-8 characters and the number of tokens in an excerpt.

To address this issue, we extend the Mistral tokenizer with 16k additional tokens specifically selected to improve character-per-token rate on Japanese, resulting in a vocabulary size of 48k tokens. The extended tokenizer significantly improves the effective context length in comparison to the Mistral 7B-v0.1 tokenizer when processing Japanese text.

By improving the tokenization process for Japanese, we can achieve cost efficient text processing during model training as well as inference.

\subsection{Foundation Model Training}
\label{ssec: training_foundation}
Previous research studies indicate that the quality of text data used for pre-training is critical to improving LLM performance \cite{rae2022scaling,dodge-etal-2021-documenting,chowdhery2022palm}. Alongside these studies, we experiment with and develop data filtering techniques that allow us to further improve the quality of available internet-scale datasets for Japanese and English. We remove any personally identifiable information in the datasets such as phone numbers, email addresses etc. 

The filtering process comprises of multiple stages, including normalization, deduplication and classification, that helps us to distinguish between high and low-quality data. This approach ensures that our model is trained on a curated dataset, enhancing its ability to generate coherent and relevant outputs. We train the released models on approximately 175 billion tokens of filtered data. 

\subsection{Model Fine-tuning}
\label{ssec: model_finetuning}
We fine-tune our foundation model to create RakutenAI-7B-instruct and RakutenAI-7B-chat using a mix of open and internally hand-crafted datasets. This tuning process enhances the model's capacity to adhere to provided instructions and prompts for RakutenAI-7B-instruct and enables RakutenAI-7B-chat to generate conversational responses akin to those of a virtual assistant, facilitating more natural dialogue interactions.

We fine-tune both the instruction- and chat-tuned models on safety datasets to mitigate the generation of explicit or offensive context, preventing the spread of misinformation or harmful biases. However, we advise caution, as they may still exhibit unintended behaviour. It is crucial to constantly monitor the model performance to ensure it adheres to ethical and social standards.  

\subsection{Evaluation}
\subsubsection{Evaluation Settings}

We evaluate the performance of our models using the Japanese \footnote{\url{https://github.com/Stability-AI/lm-evaluation-harness/tree/0fa86429679f521161d5b81a94c0c385e0a0976d }} and the English \footnote{\url{https://github.com/EleutherAI/lm-evaluation-harness/tree/b281b0921b636bc36ad05c0b0b0763bd6dd43463}}  versions of the Language Model Evaluation Harness (LM-Harness), running the evaluation with same setup to ensure fair comparison across the models. The list of the Japanese NLP tasks used for metrics is as follows:

\begin{itemize}[leftmargin=*]
    \item \textbf{JCS} / JCommonSenseQA \cite{kurihara-etal-2022-jglue} is the Japanese version of CommonSenseQA \cite{talmor-etal-2019-commonsenseqa}, which consists of multiple-choice questions that evaluate model ability in answering commonsense questions.
    \item \textbf{JNLI} / Japanese Natural Language Inference \cite{kurihara-etal-2022-jglue} is a test set to evaluate a model's ability to recognize the inference relation that a premise sentence has to a hypothesis sentence. There are three inference relations, namely: entailment, contradiction, and neutral which are presented to the model in a multiple-choice question format.
    \item \textbf{MARC-ja} / Multilingual Amazon Review Corpus is a text classification test set that is based on the MARC dataset \cite{marc_reviews}. MARC-ja is the Japanese subset of this dataset, which is a binary classification task for positive and negative review sentiment.
    \item \textbf{JSQuAD} / Japanese Stanford Question Answering Dataset \cite{kurihara-etal-2022-jglue} is a Japanese reading comprehension dataset which measures model ability to answer a question given a passage.
    \item \textbf{JAQKET} / Japanese Questions on Knowledge of Entities \cite{jaqket} is a Japanese open-domain question answering dataset which measures model knowledge on open topics such as TV quiz questions. 
    \item \textbf{XLSUM-ja} \cite{hasan-etal-2021-xl} is a Japanese subset of XLSUM, which is an abstractive summarisation dataset for 44 languages. We took the test set to measure model ability to summarise news articles. The hypothesis summary is evaluated against a reference summary using ROUGE-2 \cite{lin-2004-rouge}, a metric that measures n-gram co-occurrences.
    \item \textbf{xWino}/ xWinograd-ja is the Japanese subset of the Winograd schema challenge \cite{emelin-sennrich-2021-wino}, which is a pair of sentences that differ in only one or two contrastive words that are easily disambiguated by a human reader. This evaluation measures model's ability to use commonsense knowledge to disambiguate and debias the ambiguous sentence.
    \item \textbf{MGSM} / Multilingual Grade School Math Benchmark \cite{shi2022language} contains a set of grade-school level math problems. We took the Japanese subset to measure the models' ability to answer basic mathematical problems that require multi-step reasoning.
\end{itemize}

The list of the English NLP tasks used for metrics is as follows:
\begin{itemize}[leftmargin=*]
    \item \textbf{ARC} / AI2 Reasoning Challenge \cite{clark2018think} is a grade-school science question test set that is aimed at testing model ability to handle tough queries that basic algorithms struggle with.
    \item \textbf{HellaSwag} \cite{hellaswag} is a test set to benchmark model ability to perform commonsense reasoning, given questions that require understanding and reasoning beyond simple pattern matching.
    \item \textbf{MMLU} / Massive Multitask Language Understanding \cite{hendrycks2021measuring} is a test to measure model multitask accuracy. It covers 57 tasks that covers history, computer science, mathematics, chemistry, and other topics.
    \item \textbf{TruthfulQA} \cite{lin-etal-2022-truthfulqa} measures models' inclination to replicate common online falsehoods. 
\end{itemize}

For multiple choice question tasks, the choice with the highest likelihood is chosen as the model answer. We compute the likelihood for each choice concatenated with the question (including n-shot question-answer pairs when provided). The model answer is then compared with the reference answer to measure the accuracy.

For question-answering test sets such as JSQuAD and JAQKET, we count the model output as correct if it exactly matches the reference answer (we allow some variation in the answers by removing punctuation and emoticons, normalizing whitespaces, etc.).

In \Cref{tab:res_foundation_ja,tab:res_foundation_en}, along with the task names we indicate the metrics and the number of shots for n-shot used for evaluation. \texttt{acc} stands for the perplexity-based accuracy for multiple-choice questions, \texttt{em} stands for the exact-match metric, and \texttt{rouge-2} indicates the ROUGE-2 \cite{ganesan2018rouge} score.

\subsubsection{Evaluation Results for Foundation Models}

\begin{table*}[!h]
\centering
\small

\setlength\tabcolsep{2.5pt}
\begin{tabular}{l|c|c|cccccccc}
\toprule
\textbf{Model Name}                         & \textbf{7-Avg.} & \textbf{Avg.} & \textbf{JCS} & \textbf{JNLI } & \textbf{MARC-ja} & \textbf{JSQuAD} & \textbf{Jaqket v2} & \textbf{XLSum-ja} & \textbf{xWino} & \textbf{MGSM} \\
&{\scriptsize excl.}&&acc&acc&acc&em&em&rouge-2&acc&acc\\
&{\scriptsize XLSum-ja}&&{\scriptsize 3-shots}&{\scriptsize 3-shots}&{\scriptsize 3-shots}&{\scriptsize 2-shots}&{\scriptsize 1-shot}&{\scriptsize 1-shot}&{\scriptsize 0-shot}&{\scriptsize 5-shots} \\
\midrule
\textbf{rakuten-ai-7b}                       & \textbf{69.80}              & \textbf{62.83}            & 84.27              & 48.69               & 96.29                  & 79.09                & 80.67              & 14.08             & 77.16              & \textbf{22.40}         \\
nekomata-7b                           & 66.01              & 58.83            & \textbf{85.43}              & 40.14               & \textbf{96.80}                  & 76.29                & 71.99              & 8.59              & 73.83              & 17.60         \\
japanese-stablelm-base-gamma-7b & 64.83              & 59.12            & 80.07              & 14.71               & 92.41                  & \textbf{81.38}                & \textbf{85.05}              & \textbf{19.16}             & \textbf{82.59}              & 17.60         \\
youri-7b                              & 62.71              & 56.90            & 76.94              & \textbf{51.11}               & 90.96                  & 57.45                & 78.09              & 16.27             & 78.00              & 6.40          \\
swallow-7b                    & 60.86              & 55.18            & 78.91              & 15.16               & 90.27                  & 73.28                & 80.24              & 15.41             & 76.96              & 11.20         \\
elyza-japanese-Llama-2-7b             & 60.24              & 53.26            & 75.60              & 50.74               & 87.51                  & 71.48                & 57.56              & 4.40              & 71.22              & 7.60          \\
elyza-japanese-Llama-2-7b-fast        & 58.31              & 51.34            & 71.49              & 45.77               & 86.61                  & 70.91                & 64.18              & 2.54              & 61.63              & 7.60          \\
open-calm-7b                     & 45.27              & 39.67            & 62.65              & 31.92               & 85.37                  & 38.05                & 33.42              & 0.45              & 65.07              & 0.40         \\
\bottomrule
\end{tabular}
\caption{RakutenAI-7B foundation model performance on Japanese LM-Harness metrics in comparison with other models. Our model achieves the highest average score, more than 3 points ahead of the next best model. The models are sorted by 7-Avg.}
\label{tab:res_foundation_ja} 
\end{table*}

\begin{table*}[!h]
\centering
\small

\setlength\tabcolsep{2.5pt}
\begin{tabular}{l|c|cccc}
\toprule
\textbf{Model Name}                         & \textbf{Avg.} & \textbf{ARC}   & \textbf{HellaSwag} & \textbf{MMLU}  & \textbf{TruthfulQA} \\
& & acc & acc & acc & acc \\
& & {\scriptsize 25-shots} & {\scriptsize 10-shots} & {\scriptsize 5-shots} & {\scriptsize 6-shots} \\
\midrule
\textbf{rakuten-ai-7b}              & \textbf{60.50}   & \textbf{60.24} & \textbf{82.20}     & \textbf{61.31} & 38.25      \\
japanese-stablelm-base-gamma-7b & 56.08            & 50.60          & 77.43              & 54.99          & 41.30               \\
elyza-japanese-Llama-2-7b             & 52.76            & 51.62          & 76.54              & 44.85          & 38.02               \\
elyza-japanese-Llama-2-7b-fast        & 52.07            & 51.79          & 75.46              & 44.41          & 36.63               \\
nekomata-7b                           & 51.97            & 47.35          & 72.78              & 48.38          & 39.38               \\
youri-7b                              & 50.60            & 49.15          & 75.02              & 42.36          & 35.89               \\
swallow-7b                    & 49.90            & 47.35          & 72.20              & 39.36          & 40.68               \\
open-calm-7b                     & 29.87            & 20.56          & 31.01              & 23.73          & \textbf{44.16} \\
\bottomrule
\end{tabular}
\caption{RakutenAI-7B foundation model performance on English LM-Harness metrics in comparison with other models. Our model achieves the highest average score, more than 4 points ahead of the next best model.}
\label{tab:res_foundation_en}
\end{table*}

\Cref{tab:res_foundation_ja,tab:res_foundation_en} report the Japanese and English LM-Harness performance of RakutenAI-7B. It achieves the best overall performance compared with other Japanese 7-billion parameter foundation models on both Japanese and English test sets. The average RakutenAI-7B score for the Japanese LM-Harness evaluation is \textbf{62.83}, which is more than 3 points above Japanese-StableLM-Base-Gamma-7b, the second best model, at 59.12. We also report the average of 7 metrics {(7-Avg.)} that excludes \textbf{XLSum-ja}, as we find the performance on this task generally has a low correlation with human judgements. Our model also showcases the highest performance for 7-Avg. In general, our foundation model shows a balanced performance across all test sets, with a particularly high score on MGSM.

In addition, RakutenAI-7B retains a high performance for English, as shown by its average English LM-Harness score compared with other Japanese models. Our model achieves an average score of \textbf{60.50}, while Japanese-StableLM-Base-Gamma-7b \cite{StabilityGamma7b}, which also uses Mistral 7B-v0.1 as the base model for continual training, lags by more than 4 points compared to RakutenAI-7B, at 56.08. 

\subsubsection{Evaluation Results for Instruction-Tuned Models}
We measure the performance of the instruction-tuned models in the same manner as the foundation model.

\begin{table*}[!h]
\centering
\small
\setlength\tabcolsep{2.5pt}
\begin{tabular}{l|c|c|ccccccccc} \toprule
\textbf{Model Name}                    & \textbf{7-Avg.}    & \textbf{Avg.} & \textbf{JCS} & \textbf{JNLI } & \textbf{MARC-ja} & \textbf{JSQuAD} & \textbf{Jaqket v2} & \textbf{XLSum-ja} & \textbf{xWino} & \textbf{MGSM} \\
&{\scriptsize excl.}&&acc&acc&acc&em&em&rouge-2&acc&acc\\
&{\scriptsize XLSum-ja}&&{\scriptsize 3-shots}&{\scriptsize 3-shots}&{\scriptsize 3-shots}&{\scriptsize 2-shots}&{\scriptsize 1-shot}&{\scriptsize 1-shot}&{\scriptsize 0-shot}&{\scriptsize 5-shots} \\
\midrule
\textbf{rakuten-ai-7b-instruct}     & \textbf{77.32}    & \textbf{68.74}   & \textbf{93.03}     & \textbf{90.39}              & 96.00                  & 80.44                & 81.79              & 8.67              & 75.18              & \textbf{24.40} \\
youri-7b-instruction               & 73.35      & 66.84            & 86.06              & 70.13               & \textbf{97.03}      & \textbf{82.53}          & 79.47              & 21.29             & 79.04              & 19.20          \\
japanese-stablelm-instruct-gamma-7b   & 65.46   & 59.98            & 83.82              & 16.97               & 95.68                  & 76.20                & \textbf{81.87}     & \textbf{21.58}    & \textbf{82.06}     & 21.60          \\
swallow-7b-instruct              & 64.29       & 58.25            & 83.38              & 26.50               & 94.46                  & 75.62                & 81.01              & 16.01             & 76.23              & 12.80          \\
elyza-japanese-Llama-2-7b-instruct   & 60.04    & 53.19            & 65.15              & 57.44               & 91.51                  & 67.29                & 58.51              & 5.20              & 70.80              & 9.60           \\
elyza-japanese-Llama-2-7b-fast-instruct & 57.22 & 50.48            & 70.69              & 36.48               & 92.75                  & 68.87                & 62.29              & 3.36              & 59.44              & 10.00          \\
nekomata-7b-instruction              & 49.04   & 44.14            & 85.08              & 42.48               & 96.99                  & 8.51                 & 10.91              & 9.81              & 76.12              & 23.20         \\
\bottomrule
\end{tabular}
\caption{RakutenAI-7B-instruct model performance on Japanese LM-Harness metrics in comparison with other models. Our model achieves the highest average score, more than 3 points ahead of the next best model. The models are sorted by 7-Avg.}
\label{tab:res_instruct_ja} 
\end{table*}

\begin{table*}[!h]
\centering
\small

\setlength\tabcolsep{2.5pt}
\begin{tabular}{l|c|cccc} \toprule
\textbf{Model Name}                         & \textbf{Avg.} & \textbf{ARC}   & \textbf{HellaSwag} & \textbf{MMLU}  & \textbf{TruthfulQA} \\
& & acc & acc & acc & acc \\
& & {\scriptsize 25-shots} & {\scriptsize 10-shots} & {\scriptsize 5-shots} & {\scriptsize 6-shots} \\
\midrule
\textbf{rakuten-ai-7b-instruct} & \textbf{61.32} & \textbf{58.62} & \textbf{82.70} & \textbf{60.32} & \textbf{43.63} \\
japanese-stablelm-instruct-gamma-7b     & 55.91 & 50.43 & 77.10 & 54.61 & 41.50 \\
elyza-japanese-Llama-2-7b-fast-instruct & 54.21 & 53.58 & 77.69 & 46.91 & 38.67 \\
elyza-japanese-Llama-2-7b-instruct      & 54.07 & 52.05 & 78.33 & 47.09 & 38.83 \\
nekomata-7b-instruction                 & 52.84 & 50.34 & 73.67 & 48.53 & 38.81 \\
youri-7b-instruction                    & 52.11 & 48.98 & 75.66 & 45.41 & 38.38 \\
swallow-7b-instruct                     & 50.32 & 47.61 & 72.27 & 40.77 & 40.62 \\
\bottomrule
\end{tabular}
\caption{RakutenAI-7B-instruct model performance on English LM-Harness metrics in comparison with other models. Our model achieves the highest average score, more than 5 points ahead of the next best model.}
\label{tab:res_instruct_en}
\end{table*}

In general, we observe improvements in our instruction-tuned model over the foundation model, which shows the effectiveness of our instruction fine-tuning. \Cref{tab:res_instruct_ja,tab:res_instruct_en} report the Japanese and English LM-Harness performance of RakutenAI-7B-instruct in comparison with other Japanese instruction-tuned models within the 7 billion parameter category. The instruct model achieves an average score of \textbf{68.74}, leading by almost 2 points over Youri-7B-instruction, the second best model (this trend also holds for {7-Avg.}). Our instruct model achieves the best scores on JCS, JNLI, and MGSM. 

In addition, RakutenAI-7B-instruct improves its LM-Harness English performance compared to the foundation model, achieving the best performance as shown by its average score compared to other previously published Japanese models. It leads the next best instruct model by more than 5 points. 

\section{Conclusion}
Our systematic approach to model development, including data filtering techniques and curation, ensures that RakutenAI-7B is trained on high-quality datasets, which results in coherent and high-quality outputs for Japanese and English. The empirical evaluation results demonstrate the consistently high performance of RakutenAI-7B across a diverse range of NLP tasks, on average outperforming other open Japanese models.  Moreover, the RakutenAI-7B tokenizer is more suitable for processing Japanese text, potentially resulting in faster and cheaper training and inference.

We release the suite of RakutenAI-7B models to researchers, developers, and industry practitioners alike as part of our continuous effort to foster innovation and drive positive impact across various domains.

\section{Acknowledgements}
We want to thank Rakuten Group, Inc. for providing the resources to make this release possible.  We would also like to acknowledge the research community, and the Mistral AI team in particular, for open-sourcing and sharing their work on resource intensive foundation model research.

\section {Limitations}
The suite of RakutenAI-7B models is capable of generating human-like text on a wide range of topics. However, like all LLMs, they have limitations and can produce biased, inaccurate, or unsafe outputs. Please exercise caution and judgement while interacting with them.
\appendix

\bibliographystyle{acl_natbib}
\bibliography{main}

\end{document}